\documentclass[11pt]{article}

\usepackage{acl}

\usepackage{times}
\usepackage{latexsym}

\usepackage[T1]{fontenc}

\usepackage[utf8]{inputenc}

\usepackage{microtype}
\usepackage{newfloat}
\usepackage{listings}
\usepackage{multirow}
\usepackage{amsmath}
\usepackage{bbm}
\usepackage{subcaption}
\usepackage{graphicx}

\title{DialAug: Mixing up Dialogue Contexts in Contrastive Learning for Robust Conversational Modeling}

\author{Lahari Poddar \\ \\ \\
	\And Peiyao Wang \\ \\ Amazon \\  \texttt{\{poddarl, peiyaow, reinspac\}@amazon.com}
 \And  Julia Reinspach \\ \\ \\
}

\begin{document}
 	
\maketitle

\begin{abstract}
Retrieval-based conversational systems learn to rank response candidates for a given dialogue context by computing the similarity between their vector representations. 
However, training on a single textual form of the multi-turn context limits the ability of a model to learn representations that generalize to natural perturbations seen during inference. 
In this paper we propose a framework that incorporates augmented versions of a dialogue context into the learning objective. We utilize contrastive learning as an auxiliary objective to learn robust dialogue context representations that are invariant to perturbations injected through the augmentation method. We experiment with four benchmark dialogue datasets and demonstrate that our framework combines well with existing augmentation methods and can significantly improve over baseline BERT-based ranking architectures. 
Furthermore, we propose a novel data augmentation method, ConMix, that adds token level perturbations through stochastic mixing of tokens from other contexts in the batch. We show that our proposed augmentation method outperforms previous data augmentation approaches, and provides dialogue representations that are more robust to common perturbations seen during inference.

\end{abstract}

\section{Introduction}
Conversational systems have gained immense research attention in the past few years due to their practical applications in building intelligent digital assistants.
In order to converse with humans in natural language, a conversational system needs to produce meaningful and contextual responses at every turn of a  dialogue.
This is often accomplished by a ranking model, the goal of which is to select the most appropriate response among a set of curated candidate responses \cite{Lu2019GoalOrientedEC, Mehri2019PretrainingMF,Henderson2019TrainingNR,Xu2021LearningAE,Gu2020SpeakerAwareBF,Whang2020AnED}.
For practical applications a 
Bi-encoder model architecture is often adopted, due to its computational efficiency \cite{humeau2019poly,reimers2019sentence,TODBERT,Henderson2019TrainingNR}. In this approach, the dialogue context and candidate responses are encoded into latent vectors separately, and the ranking scores are computed based on the similarity between these vectors.

Learning effective vector representations of dialogue contexts is a challenging task. 
Since most dialogue datasets consist of free-text multi-turn interactions between humans, the \emph{exact} same context-response pair is likely to be seen only \emph{once} in the whole training set. However, 
during inference, 
the same response could be appropriate for various different forms of contexts. For example, in the customer service domain, a response such as ``I can issue a refund for the damaged item'' could be appropriate for many contexts that fall into the general theme of a customer having received a damaged item. Such contexts may differ from one another due to variations in customer language, the particular item details, or the type of damage etc. 
Hence, a response ranking model needs to learn representations that are robust to syntactic and fine-grained semantic variations in the dialogue context.

In order to learn representations with improved generalization capabilities, data augmentation
has become ubiquitous in  computer vision \cite{Shorten2019ASO}.
Recent research \cite{Shen2020ASB,Feng2021ASO,Longpre2020HowEI} has also reported success in leveraging augmentations for NLP tasks. 
An effective method of incorporating data augmentation for better representation learning is through a contrastive learning framework
\cite{Chen2020ASF,Wu2020CLEARCL,Gao2021SimCSESC,Fang2020CERTCS,Xie2020UnsupervisedDA,Fabbri2021ImprovingZA,Wei2021FewShotTC}, where the objective is to maximize similarity between encoded representations of an input and its augmented version. 
While contrastive learning with data augmentations has shown promising results in several NLP tasks, to the best of our knowledge the potential of such approaches for conversational modeling has not yet been explored.

In this work we propose a multi-objective model architecture, DialAug, for learning robust dialogue response ranking by leveraging the power of text data augmentations in combination with contrastive learning. 
During training, the model learns to predict the same response for both the original dialogue context and for its augmented version, making it agnostic to variations in the context.
To capture the notion of coherence and semantic relevance of a dialogue,
we introduce an auxiliary contrastive objective that learns the similarity between different views of a dialogue context, in contrast to views of contexts of other dialogues. 

We further propose a novel data augmentation method for \textbf{Con}text \textbf{Mix}ing (ConMix) that adds token level perturbations to the dialogue context, in order to simulate different variations of a multi-turn context.
ConMix  stochastically replaces some of the input tokens in a dialogue context with tokens from another randomly sampled context in the training batch. 
Through this method we are not only creating a perturbed version of the original context 
that will help learn generalizable representations, but are also generating hard negatives for other responses and contexts in the batch, 
due to the word overlap infused through stochastic mixing from other context in the same batch. 
To summarize, in this paper we make the following major contributions:
\begin{itemize}
	\item We propose a multi-objective model architecture, DialAug, for dialogue response ranking that uses a ranking objective and a contrastive learning objective. The proposed architecture is modular and can be effectively combined with many data augmentation techniques.
	\item We propose a novel data augmentation technique, ConMix, that stochastically adds token-level perturbations to dialogue contexts during training, leading to better performance and robustness of the learned model as compared to baseline data augmentation methods.
	\item We conduct an extensive set of evaluations on four large-scale publicly available dialogue datasets, and demonstrate the proposed approach outperforms strong baselines and is effective in learning robust representations.
\end{itemize}

\section{Related work}
We review two closely related research areas: data augmentation techniques for text data, and contrastive learning.

\textbf{Data Augmentation for Text} : Data augmentation has been widely used for computer vision tasks, in order to increase the size of a labeled dataset, and to improve robustness of the model to input noise. Typical image augmentations include cropping, flipping, rotating, resizing, applying color distortions, and Gaussian blurring \cite{Shorten2019ASO,Chen2020ASF}. 
Equivalent simple augmentation techniques have been proposed and explored for text data tasks, e.g., word deletions and permutations, and have been shown to improve the model's robustness and performance \cite{Shorten2019ASO}.
There has also been some active research into semantic augmentation techniques, such as back-translation, synonym replacement, or generative models \cite{Shorten2019ASO,Wu2020CLEARCL,Xie2020UnsupervisedDA,Kumar2019SubmodularOD,Fang2020CERTCS}. 
However, these are comparatively complex to implement, and rely on external knowledge (i.e., synonym lists) or additional models, making them only suitable for tasks where appropriate models or knowledge exists. 
In this work we only consider automatic data augmentation techniques, i.e., techniques that do not require external knowledge or additional models, and can be easily implemented for any language or task.

\textbf{Contrastive Learning} : Contrastive learning has been shown to be a powerful representation learning technique for both vision and text data tasks \cite{Chen2020ASF,Khosla2020SupervisedCL,Wu2020CLEARCL,Giorgi2020DeCLUTRDC,Gunel2021SupervisedCL}. It essentially aims to learn a better representation of the input by maximizing agreement between two similar data points. 
These data points can be either augmented versions of the same input in self-supervised learning
\cite{Chen2020ASF,Giorgi2020DeCLUTRDC,Wu2020CLEARCL}, or from the same class label
in supervised learning \cite{Gunel2021SupervisedCL,Ma2021ContrastiveFI,Khosla2020SupervisedCL}.

Contrastive learning has been explored for both pretraining and finetuning tasks in NLP. For example, \cite{Wu2020CLEARCL,Giorgi2020DeCLUTRDC,Fang2020CERTCS} use contrastive learning to pretrain large-scale transformer encoders for sentence representations, while other researchers focus on more task-specific finetuning settings, such as summarization \cite{Fabbri2021ImprovingZA}, text classification \cite{Wei2021FewShotTC},  textual similarity \cite{Gao2021SimCSESC}, and user satisfaction prediction \cite{Kachuee2020SelfSupervisedCL}. Recent work \cite{Ma2021ContrastiveFI} has demonstrated success in adopting contrastive finetuning for neural rankers in the QA domain, however, the authors do not leverage data augmentations. Our work is more similar to CLEAR \cite{Wu2020CLEARCL} which uses contrastive learning with text data augmentations for pretraining language models. 
However, we focus on the finetuning stage of dialogue response ranking and leverage augmentations for dialogue contexts in the contrastive learning objective. We use deletion and reordering based augmentations proposed in their work as baselines for ConMix.





\section{Approach}
\begin{figure}[tbp]
	\centering
	\includegraphics[width=0.95\linewidth]{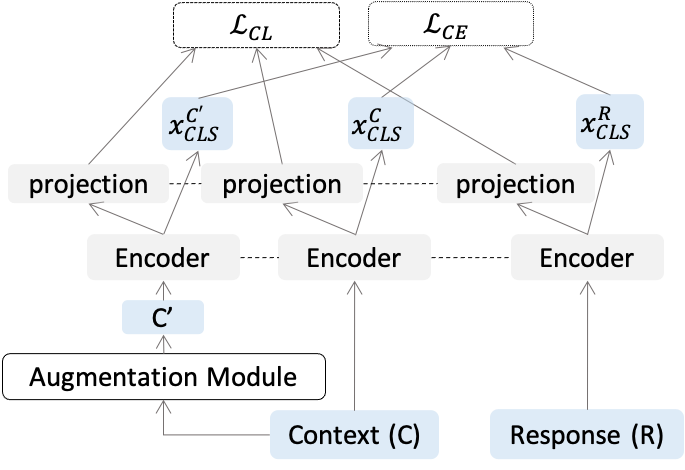}
	\caption{DialAug model architecture. Context $C$ and Response $R$ are the model inputs, and the output is a similarity score. The augmented context $C'$ and projection network are used only during training. The encoder networks share weights, as well as the projection layers.}
	\label{fig:arch}
\end{figure}

\begin{figure*}[htbp]
	\centering
	\begin{subfigure}[b]{0.37\linewidth}
		\centering
		\includegraphics[width=\linewidth]{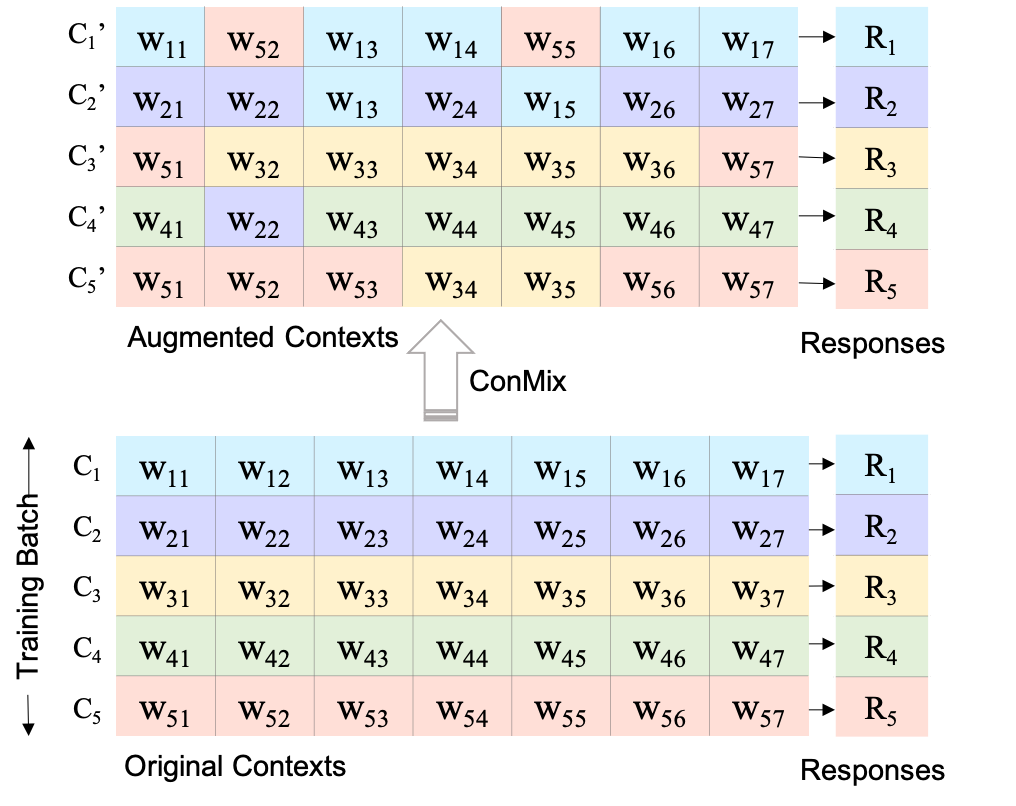}
		\caption{Mixing context tokens in batch}
		\label{fig:mixing_illus}
	\end{subfigure}
	\hfill
	\begin{subfigure}[b]{0.58\linewidth}
		\centering
		\includegraphics[width=\linewidth]{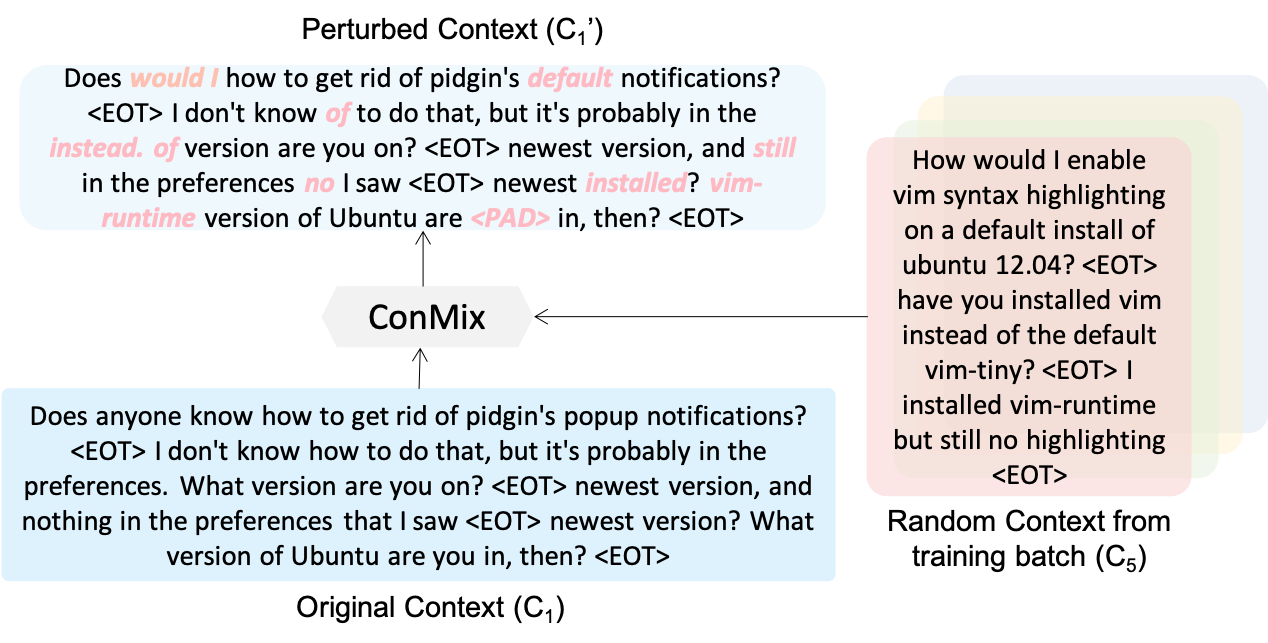}
		\caption{Example of generated perturbed context through augmentation}
		\label{fig:mixing_example}
	\end{subfigure}
	\caption{Illustration of the ConMix data augmentation. $C_1$ is a context and $R_1$ is its corresponding response. $C'_1$ is an augmented version of $C_1$, which retains most of the tokens from $C_1$ (blue), and has few tokens replaced with tokens from a random context $C_5$ (orange). $R_1$ is still considered the most appropriate response to $C'_1$.}
	\label{fig:mixing}
\end{figure*} 


Consider a batch with $B$ inputs $\{C_i, R_i\}_{i = 1, 2 , \cdots, B }$, where $C_i$ is a dialogue context and $R_i$ is the corresponding response. 
The objective of a dialogue model is to predict the most likely response $R_i$ among a set of candidate responses $\{ R_1, R_2, \cdots R_m\}$, given the dialogue context $C_i $. 

\subsection{DialAug Model Architecture}

We build upon the widely used Bi-encoder model architecture \cite{humeau2019poly,reimers2019sentence,TODBERT,Henderson2019TrainingNR}, which is efficient for real world use cases, due to its fast training and inference speed. 
Figure  \ref{fig:arch} shows the proposed model architecture.

Our architecture consists of an augmentation module that creates an additional view $C’_i$ of the dialogue context $C_i$, through certain transformations (described later). 
We first encode the context $C_i$, the augmented context $C'_i$, and the response $R_i$ to latent vectors, using a shared encoder. We leverage pre-trained language models and use BERT \cite{bert} as the encoder block. The input sequence to the BERT encoder is represented as 
\begin{equation}
C_i = [ CLS, w_1 \cdots, EOT, w_j, \cdots, w_{n-1}, EOT]
\end{equation}
where $n$ is the number of words in the context or response, $ CLS$ is a special token marking the beginning of the input sequence, and an additional end-of-turn $EOT$ token marks the end of turns in the dialogue context.
We feed these sequences to the BERT encoder and obtain latent vector representations for the input context, the augmented context, and the response.



\subsubsection{Main Task Loss}

Dialogue contexts consist of multiple turns, with a lot of information that might be redundant for predicting the next response. We argue that such lengthy contexts can usually accommodate small word level variations without changing the  overall theme or topic of the conversation and the next response.
Therefore, we consider the augmented version of a context $C'_i$ to be label-invariant. 

This allows the model to learn that $R_i$ is the next response for both the original context $C_i$ and its augmented version $C'_i$. 
Introducing these $(C'_i, R_i)$ pairs in the main task loss of response ranking essentially doubles the number of training data points seen by the model in each epoch. More importantly, this forces the model to learn robust representations of the lengthy dialogue contexts, in order to rank the response $R_i$ over other $m$ candidate responses, for two different views of it.

In order to obtain an aggregated vector representation of the sequences, we use the latent vector representation of the $CLS$ token.
The score of a candidate response $R_i$ for a context is computed using dot-product of their vector representations
\begin{align}
score(C_i, R_i) &= x^{C_i}_{CLS}\cdot x^{R_i}_{CLS}\\
score(C'_i, R_i) &= x^{C'_i}_{CLS} \cdot x^{R_i}_{ CLS}
\end{align}
where $x^{C_i}_{CLS}, x^{C'_i}_{CLS}, x^{R_i}_{ CLS}$ denote the representations from the $CLS$ token of context, the augmented context and the response, respectively. We optimize a cross-entropy loss ($\mathcal{L}_{CE}$) to achieve our main goal of scoring the next response in the dialogue higher than a set of candidate responses. During training, we consider the other responses in a batch as negatives for a given context.



\subsection{Contrastive Learning}
 
We introduce a contrastive learning objective as an auxiliary task during training. 
In particular, through the contrastive learning objective, we learn similarities between the vector representations of the original context $C_i$ and the augmented context $C'_i$.
In addition to $C_i$ and $C'_i$, the response $R_i$ is also a part of the same dialogue, and hence we include the response candidates into our contrastive loss.

Following \cite{Chen2020ASF}, we apply a projection network $g(\cdot)$ to transform the representations to a space where the contrastive loss will be applied. 
We use a simple $2$-layer feed-forward network with ReLU non-linearity.
The contrastive loss $\mathcal{L}_{CL}$ is optimized to maximize the similarity between span representations of $C$, $C’$ and $R$. 

We adopt a generalized version of the NT-Xent loss \cite{Chen2020ASF} that can accept multiple positives. For the $i^{th}$ training instance, the positive pairs are given by $\{ (z_{C_i}, z_{C'_i}), (z_{C_i}, z_{R_i}), (z_{C'_i}, z_{R_i})\}$. 
For each such positive pairs $(p_i, p^+_i)$, the contrastive loss term is represented as
\begin{equation}
	\ell_{p_i, p^+_i} = -  \text{log} \frac{exp(z_{p_i} \cdot z_{p^+_i})/ \tau}{\sum_{k = 0}^{B} \mathbbm{1} _{k \neq i}\sum_{q \in S}exp(z_{p_i} \cdot z_{q_k})/ \tau}
\end{equation}
\noindent where $\tau$ denotes the temperature in the loss, $\mathbbm{1} _{k \neq i}$ is an indicator function, $B$ is the batch size and $S = \{C, C', R\}$ are the sequences in the batch. The total contrastive loss $\mathcal{L}_{CL}$ within a batch is computed over all such positive pairs.

The overall loss is a weighted summation of the cross-entropy and the  contrastive loss, 
\begin{equation}
\mathcal{L} = \mathcal{L}_{CE} 
+ \lambda \mathcal{L}_{CL} 
\end{equation}
where $\lambda$ is a weight coefficient for the auxiliary loss. 
We empirically set this value to 0.5 for all our experiments.
A careful reader might observe that we introduce additional parameters in the skeleton Bi-encoder architecture through the projection network, however, they are only used during training and discarded afterwards. During inference, the model has a comparable number of parameters and speed as a Bi-encoder.

\subsection{ConMix Data Augmentation}
\label{MixDA}


We design a novel data augmentation method, ConMix, to generate the augmented view $C'_i$ for a given context $C_i$. ConMix creates augmentations through dynamic mixing of words from other contexts in the batch. 
In particular, for each $C_i$ it selects a random context $C_j$ from the training batch and replaces random words of $C_i$ with words in the same positions from $C_j$, to generate an augmented version ($C'_i$). Figure \ref{fig:mixing} shows an illustrative example of the mixing process.
This introduces perturbations to the original context and stochastically creates variations of it that the model learns to recognize as similar, and ranks the same response at the top among other candidates.
With the batch mixing strategy the augmented context ($C'_i$) also serves as a hard negative. This is because the augmented version $C'_i$ has significant word overlap with $C_j$, the random context from where the replacement tokens were chosen. Thus creating harder negative pairs <$C'_i, R_j$> and <$C'_i, C_j$> in the main task loss and the contrastive loss, respectively. 

We adapt the Bernoulli MixUp approach \cite{beckham2019adversarial} for mixing tokens of dialogue contexts. In $C_i'$, we wish to retain the \emph{majority} of tokens from the original context $C_i$, and replace the rest with tokens from a random context $C_j$. 
We first sample a binary mask $m \in \{0, 1\}^n$, where $n$ is the number of tokens in a context sequence. 
\begin{equation}
	C_i' = m \circ C_i + (1-m) \circ C_j, \text{ where } i \neq j
\end{equation}
$C_i'$ is the augmented view of context $C_i$, and $C_j$ is a randomly selected context from the same batch, $\circ$ denotes the Hadamard product. 
The binary mask $m$ is sampled from a $\text{Bernoulli}(\lambda_{mix})$ distribution where $\lambda_{mix} \in (0.5, 1]$ is the mixing coefficient. 
The proportion of replaced tokens is controlled by $\lambda_{mix}$. Intuitively, we should use a coefficient that retains most of the words from the original context, to ensure that the augmented context $C'_i$ is label invariant, i.e., can have the same next response $R_i$, and is more similar to $C_i$ than to $C_j$. In order to preserve the higher-level dialogue structure, we retain the end-of-turn ($EOT$) markers in the context while generating the binary mask.



The augmentations in our architecture are stochastically generated during each epoch. Therefore, for a context $C_i$, the augmented view $C_i'$ might be different across epochs, depending on the random selection of the mixing context $C_j$ and replaced token positions within $C_i$. This enables the model to see many variations of the same context and learn to generalize across these representations. 

\section{Experiments}
\subsection{Datasets}
We finetune and evaluate our response ranking models on the following four public task-oriented dialogue datasets: 

\noindent \textbf{1. Ubuntu V2: }
The Ubuntu V2 corpus \cite{lowe2015ubuntu} consists of conversations extracted from Ubuntu chat logs, where people seek technical support for various Ubuntu-related problems from the community. We use a public repository\footnote{https://github.com/rkadlec/ubuntu-ranking-dataset-creator} to generate the train/dev/test examples. 

 \noindent\textbf{2. Advising DSTC7: }
The Advising dataset from DSTC7 subtask 1 \cite{gunasekara2019dstc7} contains dialogues in which university students seek advise on classes to take. The dataset was built upon expanding 815 original conversations by paraphrasing. This dataset additionally contains profile information for students, which we do not include in our model to be consistent with other datasets. 

 \noindent \textbf{3. Ubuntu DSTC7: }
The Ubuntu DSTC7 dataset is similar to the Ubuntu V2 corpus, but it was further disentangled and annotated from the original chat logs data \cite{arxiv18disentangle}. 
We evaluate our model on the subtask 1 of the DSTC7 challenge, the goal of which is to select the most appropriate response from 100 candidates.

\noindent \textbf{4. Taskmaster: }  
This dataset consists of written dialogues in the movie ticketing domain \cite{Byrne2019Taskmaster1TA}.
We split the dialogues into train/dev/test sets\footnote{At the time of conducting our experiments, the train/dev/test splits were not yet released.},
and treat each system turn and its corresponding dialogue context as a positive pair. 
For evaluation, we randomly sample 50 negative responses per context from all available system turns. 

The dataset statistics are summarized in Table \ref{tbl: datastat}. For each dataset, we calculate the 95th percentile of its context and response lengths, and use these values as the maximum sequence length in the corresponding encoders. 
We use a batch size of 20 for Taskmaster and 32 for the other three datasets.

\subsection{Implementation Details}
We implement our models using the Pytorch deep learning framework and the HuggingFace transformer library \cite{pytorch}. For implementation of the contrastive loss we use the Pytorch metric learning library \cite{metriclearning}. We set the mixing coefficient ($\lambda_{mix}$ ) in ConMix to $0.7$, i.e., $30\%$ of the tokens are replaced.
We use {\tt bert-base-uncased} as our pre-trained encoder, and train all our models in an end-to-end manner with Adam optimizer \cite{adam} for fine-tuning. 


\begin{table}[tbp]
	\centering
	\resizebox{\linewidth}{!}{%
		\begin{tabular}{l|cccc}
			\hline
			\multirow{2}{*}{}  & \multirow{2}{*}{Ubuntu v2} & \multicolumn{2}{c}{DSTC7} & \multirow{2}{*}{Taskmaster} \\
			&                            & Ubuntu     & Advising     &                             \\ \hline
			Train examples          & 1M                         & 100k       & 100k         &             192,821                \\
			Dev examples     & 19,560                     & 5k         & 500          &             10,715                \\
			Test examples           & 18,920                     & 1k         & 500          &            10,717                 \\
			Eval candidates & 10                         & 100        & 100          &         
			51                    \\ \hline
		\end{tabular}%
	}
	\caption{Statistics of the datasets: Ubuntu V2, Ubuntu DSTC7, Advising DSTC7, and Taskmaster.}
	\label{tbl: datastat}
\end{table}

\subsection{Baseline Augmentations}\label{sec: aug}
We explore and evaluate the following augmentation methods as baselines to compare with ConMix:

\noindent\textbf{1. Subsequence sampling:} Similar to cropping \cite{Chen2020ASF} for images and span sampling \cite{Giorgi2020DeCLUTRDC} for sentences,  we explore a subsequence sampling augmentation for dialogues. We
create augmentations by randomly truncating the initial turns of a given context. We argue that a response is more closely related to later turns in the context compared to earlier ones, especially in task-oriented dialogues. Hence such a strategy can highly preserve the label from the original context.



\noindent\textbf{2. Word deletion:} We implement the word deletion augmentation and hyperparameters as described in \cite{Wu2020CLEARCL}. Following \cite{Wu2020CLEARCL}, we randomly select 70\% \footnote{We also conducted experiments with word deletion rate of 30\% similar to ConMix but it underperformed the variant with recommended 70\% deletion rate} of the words in the dialogue history and replace them with the special token $[DEL]$. We merge consecutive $[DEL]$ tokens into one.

\noindent \textbf{3. Word reordering:} We randomly sample several pairs of words and switch them pairwise. We swap 30\% of the words similar to our proposed ConMix. 
In contrast to ConMix, this method only mixes words within one dialogue context.

\noindent \textbf{4. Word replacement:} We randomly replace 30\% of the words in a context with random words. In contrast to ConMix, this method replaces context words with words from the full vocabulary, and not only with words from the same training batch. 

Similar to ConMix we protect the special token $EOT$ from being replaced in all baseline augmentations to preserve the dialogue structure.

\section{Results and Discussion}
\begin{table*}[htbp]
	\centering
	\resizebox{\linewidth}{!}{%
	\begin{tabular}{l|cc|cc|cc|cc}
		\hline
		\multirow{2}{*}{Models}    & \multicolumn{2}{c|}{Ubuntu V2} & \multicolumn{2}{c|}{Advising DSTC7} & \multicolumn{2}{c|}{Ubuntu DSTC7} & \multicolumn{2}{c}{Taskmaster} \\ \cline{2-9} 
		& R@1/10       & MRR           & R@1/100         & MRR              & R@1/100        & MRR             & R@1/50       & MRR            \\ \hline
		Bi-Encoder                                     & 82.8$\pm$.3                      & 89.5$\pm$.2                & 21.1$\pm$.4                     & 33.3$\pm$.1                & 56.7$\pm$.7                    & 66.0$\pm$.3                & 87.8$\pm$.2                     & 89.6$\pm$.2                \\
\hline \hline
		DialAug + Subsequence                         & 83.0$\pm$.0                     & 89.7$\pm$.0                & 21.6$\pm$.3                     & 34.2$\pm$.1                & 57.1$\pm$.8                     & 66.7$\pm$.7                & 86.9$\pm$.2                     & 89.1$\pm$.1                \\
		DialAug + Subsequence + CL                    & 82.9$\pm$.1                     & 89.6$\pm$.0                & 20.6$\pm$.3                     & 33.1$\pm$.3                & 56.9$\pm$.1            & 66.7$\pm$.2       & 87.6$\pm$.2                     & 89.5$\pm$.1                \\ \hline
		DialAug + Deletion                            & 83.2$\pm$.1                     & 89.8$\pm$.1                & 21.5$\pm$.1                     & 34.2$\pm$.6                & 57.4$\pm$.6                     & 67.2$\pm$.5                & 88.2$\pm$.3                     & 90.1$\pm$.2                 \\
		DialAug + Deletion + CL                       & 83.3$\pm$.1                     & 89.8$\pm$.1                & 21.7$\pm$.1                     & 34.9$\pm$.3                & \textbf{58.1}$\pm$.4         & \textbf{67.8}$\pm$.6      & 88.5$\pm$.2                     & 90.2$\pm$.1                \\ \hline
		DialAug + Reordering                          &        82.7$\pm$ .2                      &      89.5$\pm$.1         & 19.7$\pm$1.0        & 33.3$\pm$1.4     & 56.2$\pm$.6                     & 66.0$\pm$.2                & 87.9$\pm$.2                     & 89.8$\pm$.1                \\
		DialAug + Reordering + CL                     &           82.9$\pm$.1          &       89.6$\pm$.1       & 19.4$\pm$.6                     & 33.3$\pm$.2                & 55.8$\pm$.4                     & 65.5$\pm$.3                & 88.0$\pm$.1           & 89.9$\pm$.1      \\ \hline
		DialAug + Replacement                              & 82.8$\pm$.1    & 89.5$\pm$.1  & 19.4$\pm.3$ &  31.6$\pm.7$&         57.5$\pm$.7         &      67.1$\pm$.6        & 89.5$\pm$.1 &   90.9$\pm$.1   \\
		DialAug + Replacement + CL                         & 82.9$\pm$.1           & 89.6$\pm$.1       &   20.9$\pm.7$                  & 33.1$\pm.4$             &           58.0$\pm$.7          &  67.3$\pm$.4 &     89.0$\pm$.2               &      90.6$\pm$.1       \\ \hline
		DialAug + ConMix                              & 83.4$\pm$.1                     & 89.9$\pm$.0      & 21.8$\pm$1.4      & 34.9$\pm$.4    & 56.8$\pm$.3                     & 66.6$\pm$.2              & \textbf{90.4}$\pm$.1          & \textbf{91.4}$\pm$.0      \\
		DialAug + ConMix + CL                         & \textbf{83.6}$\pm$.1            & \textbf{90.0}$\pm$.0       & \textbf{23.0}$\pm$.8                     & \textbf{36.0}$\pm$.7               & 57.7$\pm$.4                     & 67.0$\pm$.1                & 90.1$\pm$.3                     & 91.3$\pm$.2              \\ \hline
			
	\end{tabular}%
}
	\caption{Results on the Ubuntu V2, Advising DSTC7, Ubuntu DSTC7, and Taskmaster datasets. Results were averaged over three runs, and $\pm$ denotes the standard deviation. The numbers in bold denote the best performing model for each dataset. }	
	\label{main_results}
\end{table*}

\begin{table*}[ht]
	\centering
	\resizebox{\linewidth}{!}{%
		\begin{tabular}{lcccccccccc}
			\hline 
			\multicolumn{11}{c}{Dataset: Ubuntu V2}                                                                                                                  \\ \hline
			\multirow{2}{*}{\begin{tabular}[c]{@{}l@{}}Augmentation \\ in training\end{tabular}}			& \multicolumn{2}{c}{truncation} & \multicolumn{2}{c}{deletion} & \multicolumn{2}{c}{reordering} & \multicolumn{2}{c}{typo} & \multicolumn{2}{c}{synonym} \\ 
			& Rec@1       & MRR            & Rec@1       & MRR         & Rec@1      & MRR           & Rec@1        & MRR           & Rec@1     & MRR        \\ \hline
			NA (Bi-encoder) & 69.0$\pm$.2 & 79.7$\pm$.1 & 69.6$\pm$.6 & 80.2$\pm$.0 & 79.6$\pm$.1 & 87.5$\pm$.0 & 80.8$\pm$.1 & 88.3$\pm$.1 & 79.6$\pm$.2 & 87.4$\pm$.1\\ 
			NA (Poly-encoder) & 69.2$\pm$.2 & 79.7$\pm$.1 & 71.1$\pm$.3 & 81.2$\pm$.2 & 80.6$\pm$.1 & 88.0$\pm$.0 & 81.9$\pm$.2 & 88.2$\pm$.1 & 80.7$\pm$.3 & 88.1$\pm$.1\\ 
			Subsequence & \textbf{72.1}$\pm$.2 & \textbf{82.1}$\pm$.2 & 68.3$\pm$.4 & 79.0$\pm$.0 & 79.8$\pm$.1 & 87.5$\pm$.1 & 81.1$\pm$.2 & 88.4$\pm$.1 & 79.5$\pm$.2 & 87.4$\pm$.1\\ 
			Deletion & 70.0$\pm$.1 & 80.3$\pm$.1 & \textbf{73.1}$\pm$.2 & \textbf{82.8}$\pm$.2 & 80.4$\pm$.1 & 87.9$\pm$.1 & 81.5$\pm$.2 & 88.7$\pm$.1 & 80.4$\pm$.1 & 87.9$\pm$.1\\ 
			Reordering & 69.4$\pm$.2 & 79.9$\pm$.1 & 72.2$\pm$.1 & 82.0$\pm$.0 & 80.5$\pm$.1 & 88.0$\pm$.0 & 81.1$\pm$.1 & 88.4$\pm$.0 & 80.5$\pm$.1 & 87.9$\pm$.0\\ 
			Replacement & 69.5$\pm.1$ & 79.7$\pm.1$ & 69.6$\pm.5$ & 80.3$\pm.3$ & 79.8$\pm.3$ & 87.5$\pm.1$ & 80.9$\pm.1$ & 88.3$\pm.1$ & 79.7$\pm.1$ & 87.5$\pm.1$\\ 
			ConMix & 68.8$\pm$.2 & 79.5$\pm$.1 & \textbf{73.1}$\pm$.1 & \textbf{82.8}$\pm$.1 & \textbf{81.3}$\pm$.1 & \textbf{88.5}$\pm$.1 & \textbf{82.1}$\pm$.1 & \textbf{89.1}$\pm$.1 & \textbf{81.2 }$\pm$.0 & \textbf{88.5}$\pm$.0 \\
			\hline
		\end{tabular}%
	}
	\resizebox{\linewidth}{!}{%
		\begin{tabular}{lcccccccccc}
			\hline
			\multicolumn{11}{c}{Dataset: Advising DSTC7}                                                                                                                  \\ \hline
			\multirow{2}{*}{\begin{tabular}[c]{@{}l@{}}Augmentation \\ in training\end{tabular}}			& \multicolumn{2}{c}{truncation} & \multicolumn{2}{c}{deletion} & \multicolumn{2}{c}{reordering} & \multicolumn{2}{c}{typo} & \multicolumn{2}{c}{synonym} \\ 
			& Rec@1       & MRR            & Rec@1       & MRR         & Rec@1      & MRR           & Rec@1        & MRR           & Rec@1     & MRR        \\ \hline
			NA (Bi-encoder) & 18.4$\pm$1.4 & 29.2$\pm$.5 & 15.5$\pm$.1 & 26.1$\pm$.1 & 14.8$\pm$.3 & 25.7$\pm$.4 & 18.3$\pm$.7 & 30.3$\pm$.4 & 19.6$\pm$.8 & 30.7$\pm$.6 \\
			NA (Poly-encoder) & 17.5$\pm$.9 & 28.5$\pm$1.4 & \textbf{17.9}$\pm$1.5 & \textbf{29.7}$\pm$1.3 & 15.0$\pm$.8 & 26.4$\pm$1.0 & 13.1$\pm$.4 & 25.1$\pm$.0 & 17.0$\pm$2.5 & 28.6$\pm$2.3 \\ 
			Subsequence & \textbf{19.7}$\pm$.1 &\textbf{31.7}$\pm$.1 & 16.3$\pm$.7 & 27.0$\pm$.4 & 15.4$\pm$.8 & 26.3$\pm$.2 & 18.9$\pm$.7 & 31.1$\pm$.5 & 18.0$\pm$.8 & 29.9$\pm$.6 \\
			Deletion & 18.6$\pm$.8 & 30.4$\pm$.1 & {17.6}$\pm$.3 & {29.4}$\pm$.2 & 16.6$\pm$.0 & 28.5$\pm$.0 & 19.3$\pm$1.3 & 32.2$\pm$.9 & 18.6$\pm$.0 & 31.8$\pm$.4 \\
			Reordering & 19.0$\pm$.0 & 29.6$\pm$.4 & 15.9$\pm$1.8 & 27.8$\pm$1.3 & 17.1$\pm$.1 & \textbf{30.4}$\pm$.5 & 18.7$\pm$1.3 & 31.9$\pm$.6 & 18.1$\pm$1.0 & 31.0$\pm$.3 \\
			Replacement & 17.7$\pm.8$ & 28.6$\pm.2$ & 14.6$\pm.5$ & 24.8$\pm.9$ & 12.7$\pm.7$ & 23.6$\pm1.1$ & 18.0$\pm.3$ & 29.7$\pm.1$ & 16.3$\pm.1$ & 28.2$\pm.1$\\
			ConMix & 18.6$\pm$.0 & 29.8$\pm$.2 & 16.2$\pm$.6 & 28.0$\pm$.1 & \textbf{18.3}$\pm$.7 & 30.2$\pm$.2 & \textbf{19.6}$\pm$1.4 & \textbf{32.7}$\pm$.5 & \textbf{20.9}$\pm$1.0 & \textbf{33.2} $\pm$.9 \\ \hline
		\end{tabular}%
	}
	\resizebox{\linewidth}{!}{%
		\begin{tabular}{lcccccccccc}
			\hline
			\multicolumn{11}{c}{Dataset: Ubuntu DSTC7}                                                                                                                  \\ \hline		
			\multirow{2}{*}{\begin{tabular}[c]{@{}l@{}}Augmentation \\ in training\end{tabular}}			& \multicolumn{2}{c}{truncation} & \multicolumn{2}{c}{deletion} & \multicolumn{2}{c}{reordering} & \multicolumn{2}{c}{typo} & \multicolumn{2}{c}{synonym} \\ 
			& Rec@1       & MRR            & Rec@1       & MRR         & Rec@1      & MRR           & Rec@1        & MRR           & Rec@1     & MRR        \\ \hline
			NA (Bi-encoder) & 42.3$\pm$.1 & 51.7$\pm$.3 & 52.3$\pm$.0 & 61.9$\pm$.3 & 47.9$\pm$.6 & 57.6$\pm$.3 & 48.0$\pm$.0 & 58.2$\pm$.4 & 51.6$\pm$.4 & 61.5$\pm$.2 \\
			NA (Poly-encoder)  & 41.2$\pm$.8 & 51.3$\pm$.3 & 49.6$\pm$1.1 & 60.1$\pm$.9 & 47.9$\pm$.5 & 57.5$\pm$.2 & 45.2$\pm$.1 & 56.2$\pm$.1 & 50.6$\pm$.4 & 61.4$\pm$.1\\ 
			Subsequence & \textbf{45.7}$\pm$.2 & \textbf{55.8}$\pm$.2 & 47.6$\pm$.6 & 58.3$\pm$.2 & 47.5$\pm$.8 & 58.3$\pm$.4 & 50.8$\pm$.3 & 61.8$\pm$.3     & 52.3$\pm$.9 & 62.6$\pm$.4\\ 
			Deletion  & 42.4$\pm$1.4 & 52.7$\pm$.6 & \textbf{54.6}$\pm$.2 & \textbf{64.3}$\pm$.3 & 50.3$\pm$.3 & 60.8$\pm$.2 & 53.5$\pm$.1 & 63.5$\pm$.1 & 52.9$\pm$.6 & 63.6$\pm$.0 \\ 
			Reordering & 41.6$\pm$.3 & 51.2$\pm$.6 & 50.1$\pm$.9 & 59.7$\pm$.7 & 52.8$\pm$.3 & 62.8$\pm$.2 & 51.5$\pm$.4 & 61.7$\pm$.3       & 51.7$\pm$.4 & 62.0$\pm$.4\\ 
			Replacement & 43.9$\pm$1  & 53.6$\pm$.9  & 49.7$\pm$1  & 59.9$\pm$1  &  52.1$\pm$2.5 & 62.3$\pm$1.7 &  54.2$\pm$.4 & \textbf{64.4}$\pm$.2  & \textbf{55.0}$\pm$.7 & \textbf{64.7}$\pm$.5  \\ 
			ConMix& 44.0$\pm$.6 & 52.8$\pm$1 & {50.5}$\pm$.4 & {60.8}$\pm$.5 & \textbf{54.1}$\pm$.5 & \textbf{63.8}$\pm$.3 & \textbf{54.5}$\pm$.4 & 64.2$\pm$.5   & {54.5}$\pm$.2 & {64.1}$\pm$.2 \\ \hline
		\end{tabular}%
	}
	\resizebox{\linewidth}{!}{%
		\begin{tabular}{lcccccccccc}
			\hline
			\multicolumn{11}{c}{Dataset: Taskmaster}                                                                                                                  \\ \hline
			\multirow{2}{*}{\begin{tabular}[c]{@{}l@{}}Augmentation \\ in training\end{tabular}}			& \multicolumn{2}{c}{truncation} & \multicolumn{2}{c}{deletion} & \multicolumn{2}{c}{reordering} & \multicolumn{2}{c}{typo} & \multicolumn{2}{c}{synonym} \\ 
			& Rec@1       & MRR            & Rec@1       & MRR         & Rec@1      & MRR           & Rec@1        & MRR           & Rec@1     & MRR        \\ \hline
			NA (Bi-encoder)                               & 76.5$\pm$.5          & 80.8$\pm$.4          & 79.5$\pm$.1          & 84.2$\pm$.1          & 76.5$\pm$.2          & 82.1$\pm$.2          & 87.6$\pm$.3          & 89.6$\pm$.2          & 84.6$\pm$.2          & 87.7$\pm$.1             \\
			NA (Poly-encoder)                             & 77.1$\pm$.0        & 81.3$\pm$.0        & 79.6$\pm$.4        & 84.2$\pm$.2        & 76.5$\pm$.3        & 82.0$\pm$.2          & 88.0$\pm$.1          & 89.8$\pm$.1  &84.8$\pm$.2& 	87.8$\pm$.1
			 \\
			Subsequence                                   & \textbf{85.7}$\pm$.2 & \textbf{88.1}$\pm$.1 & 79.9$\pm$.2          & 84.4$\pm$.2          & 74.5$\pm$.8          & 80.4$\pm$.5          & 87.6$\pm$.2          & 89.5$\pm$.1          & 84.2$\pm$.4          & 87.4$\pm$.2             \\
			Deletion                                      & 76.4$\pm$.0          & 80.6$\pm$.0          & \textbf{86.4}$\pm$.2& \textbf{88.8}$\pm$.1 & 86.0$\pm$.3         & 88.5$\pm$.2          & 88.5$\pm$.2          & 90.2$\pm$.1          & 86.5$\pm$.3          & 88.9$\pm$.2             \\
			Reordering                                    & 75.9$\pm$.3          & 80.2$\pm$.2          & 83.9$\pm$.2          & 87.2$\pm$.1          & \textbf{89.1}$\pm$.3 & \textbf{90.6}$\pm$.2 & 88.0$\pm$.1         & 89.9$\pm$.1          & 86.3$\pm$.1          & 88.8$\pm$.1             \\
			
			Replacement & 74.5$\pm$.4  & 79.5$\pm$.2  & 77.0$\pm$.5  & 82.4$\pm$.3  & 71.9$\pm$1.0  & 78.6$\pm$.6  & 86.7$\pm$.4 & 88.4$\pm$.3 & 82.3$\pm$.9 & 86.2$\pm$.6 \\ 
			
			ConMix                                        & 81.3$\pm$.4          & 81.3$\pm$.3          & 85.0$\pm$.3         & 87.9$\pm$.2          & 88.2$\pm$.3          & 90.1$\pm$.2          & \textbf{90.1}$\pm$.3 & \textbf{91.3}$\pm$.2 & \textbf{89.3}$\pm$.3 & \textbf{90.8}$\pm$.2   \\ \hline
			
		\end{tabular}%
	}
	
	\caption{
		Robustness during inference for different augmentation strategies. All models using augmentions were trained with contrastive loss. Results were averaged over three runs, and $\pm$ denotes the standard deviation.}
	\label{perturbation_results1}
\end{table*}

We use Recall@1 and MRR as evaluation metrics and report numbers after averaging over 3 runs.

\subsection{Performance on Response Ranking} 

We first demonstrate our proposed model architecture's compatibility and effectiveness with ConMix, along with other baseline data augmentations. 
For each augmentation method, we conduct an ablation study to separately understand the effects from data augmentation, and the benefits obtained from the addition of contrastive learning. 
For a fair baseline comparison we include the Bi-encoder baseline \cite{humeau2019poly}, which has a comparable number of parameters and architecture. 
Larger model architectures such as Poly-encoder \cite{humeau2019poly} or Cross-encoder \cite{wolf2019transfertransfo} are orthogonal to our approach, and can potentially be adopted as backbone architecture for our model. We leave those explorations for future work.

Results on four ranking datasets for all model variants are presented in Table \ref{main_results}. 
We observe that our proposed DialAug architecture significantly outperforms the baselines across all datasets.
Specifically, our model with proposed ConMix augmentation and contrastive loss achieves an absolute gain of $0.8\%$, $1.9\%$, $1.0\%$ and $2.3\%$
 for Recall@1 metric over Bi-encoder, on the four datasets respectively.
This shows that textual variations injected in the input sequences through augmentations
results in representations that generalize better to the unseen test set.

Second, we note that our proposed augmentation method,  ConMix, consistently outperforms the baseline augmentations in all datasets by a fair margin, except for Ubuntu DSTC7. 
We find that the word reordering augmentation, which shuffles words within a context, is not as effective as the other augmentations. In this method, words are neither introduced nor removed from the context, and the model learns from the same bag-of-words as the original context. On the other hand, through deletion augmentation words get omitted from the context, and the model needs to learn to predict the response while some words are missing. ConMix takes this a step further, and not only removes some of the words from the context, but also replaces them with other random words. This forces the model to learn the task in a much harder setting with observing many variations of the same context over the epochs. As hypothesized ConMix outperforms the global word replacement method due to the added advantage of strategic in-batch mixing, infusing word overlaps in a controlled manner and supplementing harder negatives.

Finally, we note that contrastive learning (rows with + CL) helps boost performance further, compared to corresponding model versions without the additional objective. 
This indicates the effectiveness of learning to contrast partial views of a dialogue for better representation learning of the context. 
Moreover, we see that for relatively smaller sized dataset from the DSTC7 challenge, 
contrastive learning acts as an effective regularizer and can significantly reduce standard deviations of the metrics ($1.4$ to $0.8$ for Recall@1 metric for ConMix augmentation on Advising, and $0.8$ to $0.1$  from for Subsequence augmentation on Ubuntu).

\subsection{Evaluating Robustness to Perturbations}
\label{sec:robust_exp}

Next we evaluate the robustness of data augmentation methods on various perturbations introduced in the dialogue context in the test set.
We apply the perturbations independently on the original test sets and evaluate our DialAug model architecture in combination with different training augmentation methods on these harder test sets. 
As baselines with no augmentations, apart from Bi-encoder, we also  include the more powerful Poly-encoder \cite{humeau2019poly} architecture in this evaluation. 

We introduce three perturbations that are similar to the augmentation methods used during training:

\noindent  \textbf{1. Truncation: } 
Similar to subsequence sampling, we randomly truncate dialogue contexts to remove earlier turns.

\noindent  \textbf{2. Word deletion: } 
Delete words with a $30\%$ deletion rate.

\noindent  \textbf{3. Word reordering:} 
Reorder words with $30\%$ probability.

\noindent We include two additional reformulations that are commonly observed during real-world inference:

\noindent  \textbf{4. Typos:} 
We implement the vanilla noise model \cite{namysl2020nat} with noise level $0.1$  to capture character-level variations caused by typos. We randomly change 30\% of words in the context.

\noindent  \textbf{5. Synonym replacement:} 
To capture lexical variations, we randomly replace $30\%$ of words from the context with their synonyms using a pre-defined vocabulary \cite{jia2019certified}.

As can be seen from the results of Table \ref{perturbation_results1}, training on augmented data helps significantly against adversarial examples during inference, compared to baseline models trained with no augmentation.
It is interesting to note that a more expressive model such as Poly-encoder, with an order of magnitude larger number of parameters, is still susceptible to adversarial perturbations and under-performs the proposed DialAug model that leverages data augmentations.
These experiments demonstrate that robustness to noise does not come out-of-the-box for larger models. Instead, strategic data augmentation methods such as ours, that expose a model to diverse training data, can learn to handle these variations effectively.

Comparing among different augmentation methods, it is not surprising to find that a model trained with one augmentation (e.g. subsequence sampling) performs well when exposed to that specific type of perturbations (e.g. truncation) during test. However, they do not generalize well to a different type of noise seen during test (e.g. model trained with deletion based augmentation and tested on reordering). 
ConMix, on the other hand, is consistently robust to different perturbations across the four adversarial datasets even though it had not been trained specifically for them. 
It performs on par or better than the specific data augmentations such as deletion and reordering when exposed to those perturbations during test. 
For more common and realistic variations, i.e., synonyms and typos, ConMix significantly outperforms all other methods on three datasets.
This indicates a uniformly powerful and robust representation learning through this novel augmentation strategy.

\subsection{Computational Efficiency}
ConMix is designed and implemented to generate augmentations through vectorization and therefore has the benefit of being faster to train. Tokens are randomly mixed on-the-fly within a batch to create augmentations in parallel on GPUs, through fast tensor multiplications. For many augmentation methods, such vectorization might be non-trivial and the overall speed becomes limited by the process of creating augmentations outside the training loop on much slower CPUs. For example, when training on the Taskmaster dataset with 8 gpus, the DialAug architecture with ConMix is $1.2$x  faster than training with the global word replacement augmentation. While conducting full training over $20$ epochs this leads to an overall speed up by $1.5$ hours for training with the ConMix augmentation.


\section{Summary}

In this work we proposed DialAug, a modular architecture for conversational response ranking. It combines the traditional cross-entropy loss for ranking with a contrastive counterpart to learn from augmented views of the dialogue context. We presented a novel data augmentation method, ConMix, which
generates multiple views of the same context via stochastic mixing of tokens from other contexts in the batch during training. 
We conducted an extensive set of experiments on four datasets and show that a model trained with ConMix outperforms strong baselines and other augmentation methods. Our proposed model is also proven to be robust against common perturbations encountered during inference. We hope our work encourages further research in such data-centric methods to improve robustness of NLP models for practical applications of conversational modeling.


\bibliographystyle{acl_natbib}
\bibliography{references}

\begin{thebibliography}{38}
\expandafter\ifx\csname natexlab\endcsname\relax\def\natexlab#1{#1}\fi

\bibitem[{Beckham et~al.(2019)Beckham, Honari, Verma, Lamb, Ghadiri, Hjelm,
  Bengio, and Pal}]{beckham2019adversarial}
Christopher Beckham, Sina Honari, Vikas Verma, Alex Lamb, Farnoosh Ghadiri,
  R~Devon Hjelm, Yoshua Bengio, and Christopher Pal. 2019.
\newblock On adversarial mixup resynthesis.
\newblock \emph{arXiv preprint arXiv:1903.02709}.

\bibitem[{Byrne et~al.(2019)Byrne, Krishnamoorthi, Sankar, Neelakantan,
  Duckworth, Yavuz, Goodrich, Dubey, Cedilnik, and
  Kim}]{Byrne2019Taskmaster1TA}
B.~Byrne, K.~Krishnamoorthi, Chinnadhurai Sankar, Arvind Neelakantan, Daniel
  Duckworth, Semih Yavuz, Ben Goodrich, Amit Dubey, A.~Cedilnik, and Kyu-Young
  Kim. 2019.
\newblock Taskmaster-1: Toward a realistic and diverse dialog dataset.
\newblock In \emph{EMNLP/IJCNLP}.

\bibitem[{Chen et~al.(2020)Chen, Kornblith, Norouzi, and Hinton}]{Chen2020ASF}
Ting Chen, Simon Kornblith, Mohammad Norouzi, and Geoffrey Hinton. 2020.
\newblock A simple framework for contrastive learning of visual
  representations.
\newblock In \emph{International conference on machine learning}, pages
  1597--1607. PMLR.

\bibitem[{Devlin et~al.(2019)Devlin, Chang, Lee, and Toutanova}]{bert}
J.~Devlin, Ming-Wei Chang, Kenton Lee, and Kristina Toutanova. 2019.
\newblock Bert: Pre-training of deep bidirectional transformers for language
  understanding.
\newblock In \emph{NAACL}.

\bibitem[{Fabbri et~al.(2021)Fabbri, Han, Li, Li, Ghazvininejad, Joty, Radev,
  and Mehdad}]{Fabbri2021ImprovingZA}
Alexander~Richard Fabbri, Simeng Han, Haoyuan Li, Haoran Li, Marjan
  Ghazvininejad, Shafiq Joty, Dragomir Radev, and Yashar Mehdad. 2021.
\newblock Improving zero and few-shot abstractive summarization with
  intermediate fine-tuning and data augmentation.
\newblock In \emph{Proceedings of the 2021 Conference of the North American
  Chapter of the Association for Computational Linguistics: Human Language
  Technologies}, pages 704--717.

\bibitem[{Fang and Xie(2020)}]{Fang2020CERTCS}
Hongchao Fang and P.~Xie. 2020.
\newblock Cert: Contrastive self-supervised learning for language
  understanding.
\newblock \emph{ArXiv}, abs/2005.12766.

\bibitem[{Feng et~al.(2021)Feng, Gangal, Wei, Chandar, Vosoughi, Mitamura, and
  Hovy}]{Feng2021ASO}
Steven~Y. Feng, Varun Gangal, Jason Wei, Sarath Chandar, Soroush Vosoughi,
  T.~Mitamura, and E.~Hovy. 2021.
\newblock A survey of data augmentation approaches for nlp.
\newblock \emph{ArXiv}, abs/2105.03075.

\bibitem[{Gao et~al.(2021)Gao, Yao, and Chen}]{Gao2021SimCSESC}
Tianyu Gao, Xingcheng Yao, and Danqi Chen. 2021.
\newblock Simcse: Simple contrastive learning of sentence embeddings.
\newblock \emph{ArXiv}, abs/2104.08821.

\bibitem[{Giorgi et~al.(2020)Giorgi, Nitski, Bader, and
  Wang}]{Giorgi2020DeCLUTRDC}
John~Michael Giorgi, Osvald Nitski, Gary~D Bader, and Bo~Wang. 2020.
\newblock Declutr: Deep contrastive learning for unsupervised textual
  representations.
\newblock \emph{ArXiv}, abs/2006.03659.

\bibitem[{Gu et~al.(2020)Gu, Li, Liu, Zhu, Ling, Su, and
  Wei}]{Gu2020SpeakerAwareBF}
Jia-Chen Gu, Tianda Li, Quan Liu, Xiao-Dan Zhu, Zhenhua Ling, Zhiming Su, and
  Si~Wei. 2020.
\newblock Speaker-aware bert for multi-turn response selection in
  retrieval-based chatbots.
\newblock \emph{Proceedings of the 29th ACM International Conference on
  Information \& Knowledge Management}.

\bibitem[{Gunasekara et~al.(2019)Gunasekara, Kummerfeld, Polymenakos, and
  Lasecki}]{gunasekara2019dstc7}
Chulaka Gunasekara, Jonathan~K Kummerfeld, Lazaros Polymenakos, and Walter
  Lasecki. 2019.
\newblock Dstc7 task 1: Noetic end-to-end response selection.
\newblock In \emph{Proceedings of the First Workshop on NLP for Conversational
  AI}, pages 60--67.

\bibitem[{Gunel et~al.(2020)Gunel, Du, Conneau, and
  Stoyanov}]{Gunel2021SupervisedCL}
Beliz Gunel, Jingfei Du, Alexis Conneau, and Veselin Stoyanov. 2020.
\newblock Supervised contrastive learning for pre-trained language model
  fine-tuning.
\newblock In \emph{International Conference on Learning Representations}.

\bibitem[{Henderson et~al.(2019)Henderson, Vulic, Gerz, Casanueva,
  Budzianowski, Coope, Spithourakis, Wen, Mrksic, and hao
  Su}]{Henderson2019TrainingNR}
Matthew Henderson, Ivan Vulic, D.~Gerz, I.~Casanueva, Paweł Budzianowski, Sam
  Coope, Georgios~P. Spithourakis, Tsung-Hsien Wen, N.~Mrksic, and Pei hao Su.
  2019.
\newblock Training neural response selection for task-oriented dialogue
  systems.
\newblock In \emph{ACL}.

\bibitem[{Humeau et~al.(2019)Humeau, Shuster, Lachaux, and
  Weston}]{humeau2019poly}
Samuel Humeau, Kurt Shuster, Marie-Anne Lachaux, and Jason Weston. 2019.
\newblock Poly-encoders: Architectures and pre-training strategies for fast and
  accurate multi-sentence scoring.
\newblock In \emph{International Conference on Learning Representations}.

\bibitem[{Jia et~al.(2019)Jia, Raghunathan, G{\"o}ksel, and
  Liang}]{jia2019certified}
Robin Jia, Aditi Raghunathan, Kerem G{\"o}ksel, and Percy Liang. 2019.
\newblock Certified robustness to adversarial word substitutions.
\newblock In \emph{Proceedings of the 2019 Conference on Empirical Methods in
  Natural Language Processing and the 9th International Joint Conference on
  Natural Language Processing (EMNLP-IJCNLP)}, pages 4129--4142.

\bibitem[{Kachuee et~al.(2021)Kachuee, Yuan, Kim, and
  Lee}]{Kachuee2020SelfSupervisedCL}
Mohammad Kachuee, Hao Yuan, Young-Bum Kim, and Sungjin Lee. 2021.
\newblock Self-supervised contrastive learning for efficient user satisfaction
  prediction in conversational agents.
\newblock In \emph{Proceedings of the 2021 Conference of the North American
  Chapter of the Association for Computational Linguistics: Human Language
  Technologies}, pages 4053--4064.

\bibitem[{Khosla et~al.(2020)Khosla, Teterwak, Wang, Sarna, Tian, Isola,
  Maschinot, Liu, and Krishnan}]{Khosla2020SupervisedCL}
Prannay Khosla, Piotr Teterwak, Chen Wang, Aaron Sarna, Yonglong Tian, Phillip
  Isola, Aaron Maschinot, Ce~Liu, and Dilip Krishnan. 2020.
\newblock Supervised contrastive learning.
\newblock \emph{Advances in Neural Information Processing Systems}, 33.

\bibitem[{Kingma and Ba(2015)}]{adam}
Diederik~P. Kingma and Jimmy Ba. 2015.
\newblock Adam: A method for stochastic optimization.
\newblock \emph{CoRR}, abs/1412.6980.

\bibitem[{Kumar et~al.(2019)Kumar, Bhattamishra, Bhandari, and
  Talukdar}]{Kumar2019SubmodularOD}
Ashutosh Kumar, S.~Bhattamishra, Manik Bhandari, and P.~Talukdar. 2019.
\newblock Submodular optimization-based diverse paraphrasing and its
  effectiveness in data augmentation.
\newblock In \emph{NAACL-HLT}.

\bibitem[{Kummerfeld et~al.(2018)Kummerfeld, Gouravajhala, Peper, Athreya,
  Gunasekara, Ganhotra, Patel, Polymenakos, and Lasecki}]{arxiv18disentangle}
Jonathan~K. Kummerfeld, S.~R. Gouravajhala, Joseph Peper, V.~Athreya,
  R.~Chulaka Gunasekara, Jatin Ganhotra, S.~Patel, L.~Polymenakos, and
  Walter~S. Lasecki. 2018.
\newblock Analyzing assumptions in conversation disentanglement research
  through the lens of a new dataset and model.
\newblock \emph{ArXiv}, abs/1810.11118.

\bibitem[{Longpre et~al.(2020)Longpre, Wang, and DuBois}]{Longpre2020HowEI}
Shayne Longpre, Yu~Wang, and Chris DuBois. 2020.
\newblock How effective is task-agnostic data augmentation for pretrained
  transformers?
\newblock In \emph{Proceedings of the 2020 Conference on Empirical Methods in
  Natural Language Processing: Findings}, pages 4401--4411.

\bibitem[{Lowe et~al.(2015)Lowe, Pow, Serban, and Pineau}]{lowe2015ubuntu}
Ryan Lowe, Nissan Pow, Iulian~Vlad Serban, and Joelle Pineau. 2015.
\newblock The ubuntu dialogue corpus: A large dataset for research in
  unstructured multi-turn dialogue systems.
\newblock In \emph{Proceedings of the 16th Annual Meeting of the Special
  Interest Group on Discourse and Dialogue}, pages 285--294.

\bibitem[{Lu et~al.(2019)Lu, Srivastava, Kramer, Elfardy, Kahn, Wang, and
  Bhardwaj}]{Lu2019GoalOrientedEC}
Y.~Lu, Manisha Srivastava, Jared Kramer, Heba Elfardy, Andrea Kahn, Song Wang,
  and Vikas Bhardwaj. 2019.
\newblock Goal-oriented end-to-end conversational models with profile features
  in a real-world setting.
\newblock In \emph{NAACL}.

\bibitem[{Ma et~al.(2021)Ma, Santos, and Arnold}]{Ma2021ContrastiveFI}
Xiaofei Ma, C.~D. Santos, and Andrew~O. Arnold. 2021.
\newblock Contrastive fine-tuning improves robustness for neural rankers.
\newblock In \emph{EMNLP FINDINGS}.

\bibitem[{Mehri et~al.(2019)Mehri, Razumovskaia, Zhao, and
  Esk{\'e}nazi}]{Mehri2019PretrainingMF}
Shikib Mehri, E.~Razumovskaia, Tiancheng Zhao, and M.~Esk{\'e}nazi. 2019.
\newblock Pretraining methods for dialog context representation learning.
\newblock In \emph{ACL}.

\bibitem[{Musgrave et~al.(2020)Musgrave, Belongie, and Lim}]{metriclearning}
Kevin Musgrave, Serge Belongie, and Ser-Nam Lim. 2020.
\newblock \href {http://arxiv.org/abs/2008.09164} {Pytorch metric learning}.

\bibitem[{Namysl et~al.(2020)Namysl, Behnke, and K{\"o}hler}]{namysl2020nat}
Marcin Namysl, Sven Behnke, and Joachim K{\"o}hler. 2020.
\newblock Nat: Noise-aware training for robust neural sequence labeling.
\newblock In \emph{Proceedings of the 58th Annual Meeting of the Association
  for Computational Linguistics}, pages 1501--1517.

\bibitem[{Reimers and Gurevych(2019)}]{reimers2019sentence}
Nils Reimers and Iryna Gurevych. 2019.
\newblock Sentence-bert: Sentence embeddings using siamese bert-networks.
\newblock In \emph{Proceedings of the 2019 Conference on Empirical Methods in
  Natural Language Processing and the 9th International Joint Conference on
  Natural Language Processing (EMNLP-IJCNLP)}, pages 3982--3992.

\bibitem[{Shen et~al.(2020)Shen, Zheng, Shen, Qu, and Chen}]{Shen2020ASB}
Dinghan Shen, Ming Zheng, Yelong Shen, Yanru Qu, and Weizhu Chen. 2020.
\newblock A simple but tough-to-beat data augmentation approach for natural
  language understanding and generation.
\newblock \emph{ArXiv}, abs/2009.13818.

\bibitem[{Shorten and Khoshgoftaar(2019)}]{Shorten2019ASO}
Connor Shorten and T.~Khoshgoftaar. 2019.
\newblock A survey on image data augmentation for deep learning.
\newblock \emph{Journal of Big Data}, 6:1--48.

\bibitem[{Wei et~al.(2021)Wei, Huang, Vosoughi, Cheng, and
  Xu}]{Wei2021FewShotTC}
Jason Wei, Chengyu Huang, Soroush Vosoughi, Yu~Cheng, and Shiqi Xu. 2021.
\newblock Few-shot text classification with triplet networks, data
  augmentation, and curriculum learning.
\newblock In \emph{Proceedings of the 2021 Conference of the North American
  Chapter of the Association for Computational Linguistics: Human Language
  Technologies}, pages 5493--5500.

\bibitem[{Whang et~al.(2020)Whang, Lee, Lee, Yang, Oh, and Lim}]{Whang2020AnED}
Taesun Whang, Dongyub Lee, Chanhee Lee, Kisu Yang, Dongsuk Oh, and Heuiseok
  Lim. 2020.
\newblock An effective domain adaptive post-training method for bert in
  response selection.
\newblock In \emph{INTERSPEECH}.

\bibitem[{Wolf et~al.(2020)Wolf, Debut, Sanh, Chaumond, Delangue, Moi, Cistac,
  Rault, Louf, Funtowicz, and Brew}]{pytorch}
Thomas Wolf, Lysandre Debut, Victor Sanh, Julien Chaumond, Clement Delangue,
  Anthony Moi, Pierric Cistac, Tim Rault, R'emi Louf, Morgan Funtowicz, and
  Jamie Brew. 2020.
\newblock Transformers: State-of-the-art natural language processing.
\newblock In \emph{EMNLP}.

\bibitem[{Wolf et~al.(2019)Wolf, Sanh, Chaumond, and
  Delangue}]{wolf2019transfertransfo}
Thomas Wolf, Victor Sanh, Julien Chaumond, and Clement Delangue. 2019.
\newblock \href {http://arxiv.org/abs/1901.08149} {Transfertransfo: {A}
  transfer learning approach for neural network based conversational agents}.
\newblock \emph{CoRR}, abs/1901.08149.

\bibitem[{Wu et~al.(2020{\natexlab{a}})Wu, Hoi, Socher, and Xiong}]{TODBERT}
Chien-Sheng Wu, S.~Hoi, R.~Socher, and Caiming Xiong. 2020{\natexlab{a}}.
\newblock Tod-bert: Pre-trained natural language understanding for
  task-oriented dialogue.
\newblock In \emph{EMNLP}.

\bibitem[{Wu et~al.(2020{\natexlab{b}})Wu, Wang, Gu, Khabsa, Sun, and
  Ma}]{Wu2020CLEARCL}
Z.~Wu, Sinong Wang, Jiatao Gu, Madian Khabsa, Fei Sun, and Hao Ma.
  2020{\natexlab{b}}.
\newblock Clear: Contrastive learning for sentence representation.
\newblock \emph{ArXiv}, abs/2012.15466.

\bibitem[{Xie et~al.(2020)Xie, Dai, Hovy, Luong, and
  Le}]{Xie2020UnsupervisedDA}
Qizhe Xie, Zihang Dai, Eduard Hovy, Thang Luong, and Quoc Le. 2020.
\newblock Unsupervised data augmentation for consistency training.
\newblock \emph{Advances in Neural Information Processing Systems}, 33.

\bibitem[{Xu et~al.(2021)Xu, Tao, Jiang, Zhao, Zhao, and
  Yan}]{Xu2021LearningAE}
Ruijian Xu, Chongyang Tao, Daxin Jiang, Xueliang Zhao, Dongyan Zhao, and Rui
  Yan. 2021.
\newblock Learning an effective context-response matching model with
  self-supervised tasks for retrieval-based dialogues.
\newblock In \emph{AAAI}.

\end{thebibliography}

\end{document}